\pdfoutput=1

\documentclass[11pt]{article}

\usepackage[preprint]{acl}

\usepackage{times}
\usepackage{latexsym}
\usepackage{enumitem}
\usepackage[T1]{fontenc}
\usepackage[utf8]{inputenc}

\usepackage{microtype}

\usepackage{inconsolata}
\usepackage{amsmath}
\usepackage{pifont}
\usepackage{colortbl}
\usepackage{subfig}
\usepackage{cleveref}
\usepackage{booktabs}
\usepackage{multirow}
\usepackage{todonotes}
\usepackage{amsthm}
\usepackage{tablefootnote}
\usepackage{bbding}
\usepackage{pifont}
\usepackage{wasysym}
\usepackage{amssymb}

\newcommand{\xmark}{\ding{55}}
\title{
Beyond Accuracy Optimization: \\ Computer Vision Losses for Large Language Model Fine-Tuning
}

\author{Daniele Rege Cambrin, Giuseppe Gallipoli, Irene Benedetto \\ \textbf{Luca Cagliero, Paolo Garza} \\
  Politecnico di Torino \\
  \texttt{\{daniele.regecambrin,giuseppe.gallipoli,irene.benedetto\}@polito.it} \\
  \texttt{\{luca.cagliero,paolo.garza\}@polito.it}
}

\begin{document}
\maketitle

\begin{abstract}
Large Language Models (LLMs) have demonstrated impressive performance across various tasks. However, current training approaches combine standard cross-entropy loss with extensive data, human feedback, or ad hoc methods to enhance performance. These solutions are often not scalable or feasible due to their associated costs, complexity, or resource requirements. This study investigates the use of established semantic segmentation loss functions in natural language generation to create a versatile, practical, and scalable solution for fine-tuning different architectures. 
We evaluate their effectiveness in solving Math Word Problems and question answering across different models of varying sizes. For the analyzed tasks, we found that the traditional Cross-Entropy loss represents a sub-optimal choice, while models trained to minimize alternative (task-dependent) losses, such as Focal or Lovász, achieve a mean improvement of +42\% on exact match without requiring additional data or human feedback. These findings suggest a promising pathway for more efficient and accessible training processes.

\end{abstract}

\section{Introduction}

\begin{figure}[htb]
    \centering
    \includegraphics[width=\linewidth]{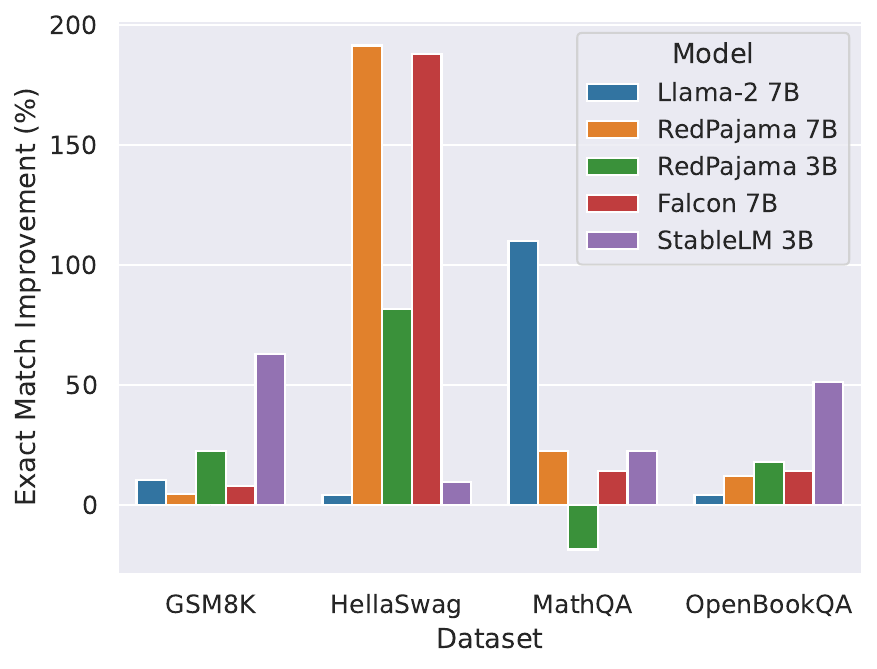}
    \caption{Percentage of improvement using the best loss (among Focal, Lovász, Generalized Dice, and Self-Adjusting Dice) for the task with Cross-Entropy compared to model fine-tuned with Cross-Entropy only.}
    \label{fig:visual_abstract}
\end{figure}

Generative Language Models have shown impressive capabilities across various scenarios \cite{t5}. Recent advancements in Large Language Models have made this even more evident \cite{helm}. However, the performance of these models is influenced by three main factors: model size, amount of training data, and training strategy \cite{wizardmath,mammoth}. Increasing model size requires more computational resources while training on vast data collections is essential to achieve competitive results when increasing the size. Additional training refinements have been introduced recently, some of which involve human experts providing feedback to enhance model performance, as in Reinforcement Learning from Human Feedback (RLHF) \cite{rlhf} where human preferences are then used to align the model outputs.

Despite the improved performance, developing these models requires massive amounts of resources, power, time, and therefore significant costs, making them accessible only to very few leading companies.
The increasing costs and resource requirements have already led to the development of solutions aimed at democratizing the training of these models. An example of this is the use of Parameter Efficient Fine-Tuning (PEFT) \cite{peft} like Low-Rank Adaptation (LoRA) \cite{lora} or derived strategies, often combined with quantization techniques, to enable lightweight fine-tuning of these computationally expensive architectures. Some works (e.g., Zephyr \cite{zephyr}) circumvent the need for human feedback by distilling the knowledge of larger and more powerful models (e.g., GPT-4), leveraging the so-called AI Feedback (AIF). However, this may cause the propagation of potential biases from the larger model, or the generation of unfactual content, potentially resulting in a less accurate representation of reality \cite{liu2023perspectives, hosking2024human}.
Although Direct Preference Optimization (DPO)~\cite{dpo} and similar methods address the instability in reinforcement learning training \cite{dpo,eisl_loss} and the computational requirements, their effectiveness is limited by the quality of the prior supervised fine-tuning stage \cite{cpo,zephyr}. 

Supervised fine-tuning still remains the most memory-efficient alternative, however
the collection of fine-tuning and instruction-tuning datasets presents similar challenges, as it requires additional annotation costs that may not always be accessible, especially when dealing with multiple datasets to annotate \cite{mammoth}. While costs can be mitigated by relying on weak annotation, similar to possible issues of using AIF, the quality of annotations is not always guaranteed.
The need for larger models and additional data pertains not only to general-purpose systems (e.g., ChatGPT) but also to task-specific models for more complex scenarios, as in the case of Math Word Problems (MWP). Specifically, state-of-the-art solutions often employ more complex training procedures, involving multiple training stages (e.g., supervised fine-tuning, instruction tuning, preference-based tuning) \cite{wizardmath, mammoth, llemma} to achieve improved performance.
However, similar to general-purpose models, this leads to more expensive solutions. Additionally, these approaches are also less portable since they are tailored to one specific task.

To tackle the above-mentioned challenges, in this work we raise some concerns about the standard practice of cross-entropy (CE) loss optimization, which is the usual language modeling objective, and we show that a more accurate selection of the loss function to optimize can be incredibly beneficial for model training.
Specifically, by using a loss function tailored to the task under analysis and leveraging LoRA, we effectively fine-tune LLMs with small amounts of data. 
The underlying idea is that for certain language tasks, it is desirable to optimize not only for the correctness of the output but also for the structural adherence of the generated text to a specific format or syntax. This is particularly relevant for tasks involving formal languages or well-defined procedures, such as mathematical reasoning, where the intermediate steps and reasoning process must follow a strict structure. Consequently, we hypothesize that accounting for these characteristics by employing a more appropriate loss function could lead to improved performance.

Our solution neither involves the implementation of complex training procedures (e.g., \cite{wizardmath}) nor further pre-training (e.g., \cite{llemma}), nor distilling knowledge from larger models (e.g., \cite{zephyr}) nor collecting human preferences \cite{OpenAIGPT42023}. In contrast, our approach focuses on selecting a more suitable loss function based on the task at hand, achieving improvements over the standard cross-entropy loss through a single training stage, as shown in \Cref{fig:visual_abstract}. In this work, we focus on mathematical reasoning and closed-question answering, which are common benchmark tasks \cite{helm,bigbench,mmlu} that have a clear and well-defined expected output.
Furthermore, we choose these tasks since we claim that, for both of them, the role of human preferences is secondary since it is difficult to ``prefer'' one output with respect to another, especially in mathematical reasoning where there could be more than one procedure to get the final solution.

Our results show that accurately choosing the right loss function (combined with Cross-Entropy) can improve performance on the analyzed tasks using the same amount of data without adding complexity to the training process. 

The source code to reproduce the experiments is available for research purposes at \url{https://github.com/DarthReca/segmentation-losses-nlp}.

\section{Related Works}

The evolution of Large Language Models has been driven by various innovative training methods. This section provides an overview of the existing approaches for training LLMs, highlighting the challenges and benefits of each approach. Additionally, we explore the development of alternative loss functions beyond cross-entropy in both natural language processing and computer vision fields.

\subsection{Training methods for Large Language Models}
The most common approaches to training LLMs are pre-training, instruction tuning, supervised fine-tuning, and tuning by preferences.

\paragraph{Pre-Training.}
Among these, effective pre-training remains a key solution for achieving the best results \cite{llemma,mistral,lima}. However, the need for high computational resources and a large amount of usable data (e.g., the source license must grant permission for the intended scope) makes this solution not always scalable or feasible in most cases. 

\paragraph{Supervised Fine-tuning and Instruction Tuning.}
Supervised fine-tuning and instruction tuning are common solutions to adapt pre-trained models to a series of tasks \cite{wizardlm,mammoth,mistral}, as this approach requires less data and can exploit efficient solutions like LoRA \cite{lora} and quantization to scale the training. However, there is still a need for large data collections since the employed language modeling loss (commonly Cross-Entropy) does not effectively represent the salient parts of the instructions (e.g., it may not correctly represent the token distribution \cite{focal_loss}). In many cases, ad hoc collections must be created, and since costs and time are still high, many solutions leverage other language models to create annotations \cite{mammoth,openorca,metamath}. Although this is a more cost-effective solution than human annotation, it could lead to biased datasets \cite{llm_annotation_survey}.

\paragraph{Human Feedback.}
RLHF \cite{rlhf} and DPO \cite{dpo} propose new methods to train models using human preferences.

Human feedback has proven helpful in tasks that require evaluating the model's text generation capability \cite{human_feedback_summarization,eli5,standford_preference_dataset}, where quantitative evaluation alone cannot cover all aspects of the desired output \cite{eval_metrics_survey}. However,~\citet{lima} highlights the limitations of RLHF, while ~\citet{hosking2024human} argues that preference scores under-represent crucial aspects such as factuality, which is an objective for question-answering and mathematical reasoning.

Moreover, this approach requires collecting human preferences, which is costly. In this case, some solutions use distilled feedback to avoid extra costs \cite{zephyr}, although this exposes them to potential biases of the employed model.

WizardMath \cite{wizardmath} proposes a different approach to include feedback in mathematical problems named Reinforcement Learning from Evol-Instruct Feedback (RLEIF). Although human feedback (with its potential biases) is avoided, RLEIF faces scalability issues due to the need for training two additional models (i.e., Instruction Reward and Process-Supervised Reward models) to produce various feedback types.

\subsection{Loss functions beyond Cross-Entropy}

Despite the most common approaches involving cross-entropy and the optimization of feedback through RL, other methods exist. Reinforcement learning has already been used to optimize the BLEU metric~\cite{Ranzato2015SequenceLT,DBLP:journals/corr/abs-1902-10245}, before the employment of feedback. However, the training instability and the unclear contribution in some settings \cite{rl_machine_traslation} are great drawbacks, coupled with its non-differentiability. EISL loss~\cite{eisl_loss} was proposed to optimize the n-grams matching in a differentiable and more stable way, but it is applicable to non-autoregressive models. Self-Adjusting Dice Loss \cite{dice_nlp}, a combination of Dice and Focal losses, was proposed to address imbalanced classification tasks in NLP. However, they employed encoder-only architectures, and the benefits depend on the specific task.
Dice and Focal losses originate from the computer vision field (in particular semantic segmentation), which is rich in loss functions designed to address class imbalance (which translates to token imbalance in NLG) and effectively penalize prediction errors. Dice \cite{dice_loss}, Generalized Dice \cite{gdice_loss}, Focal \cite{focal_loss}, and Lovász \cite{lovasz_loss} are some established loss functions that aim to address these issues by optimizing objectives other than accuracy (e.g., Dice score, Intersection-over-Union), unlike Cross-Entropy \cite{dice_nlp}. Additionally, their combination has proven to be more effective than employing them singularly in computer vision \cite{combo_loss,unified_focal,dice_focal,topologyaware}. 

Transferring this approach to causal language modeling is particularly appealing since these loss functions are differentiable, stable during training, and generalizable to many tasks. Although existing solutions for causal language modeling have tried to improve the training in different ways, they still suffer from scalability problems in terms of data, training time, and costs. This work aims to propose a simple, generalizable, and scalable approach to improve existing models without involving large data collection or complex training strategies. We show that a better extraction of knowledge from existing data can already provide relevant results' improvements by training only a few parameters (using LoRA) and small data collections (between 500 and 40K samples).

\section{Methodology}
\label{sec:methodology}

In this section, we formally introduce the loss functions we employ, shortlisted from the classification presented in~\citet{segmentation_loss_odyssey}, 
and explain their rationale. %
We describe our approach when employing them for causal generation.
For the sake of readability, all loss formulations are reported in \Cref{sec:loss_formulation}.

\subsection{Distribution-based losses}
This family of loss functions 
is derived from the Kullback-Leibler Divergence. They aim to 
optimize the model weights according to the differences between the observed and expected distributions.

\paragraph{Cross Entropy Loss.}
Cross-Entropy (CE) is an accuracy-oriented function, i.e., it aims  
to maximize the accuracy (AC) metric globally in the predicted tokens~\cite{dice_nlp}. 
CE is the most established loss for pre-training and fine-tuning language models.
Cross-entropy does not consider the underlying structures of predictions or any differences between classes and errors. Class imbalance is common in language problems, where classes are represented by tokens in the vocabulary, and token distributions are rather diversified (see \Cref{sec:token_distribution}). 
Although weighted cross-entropy may address this issue, assigning a proper weight to each class (i.e., token) can be challenging.

\paragraph{Focal Loss.}
Focal Loss (FL) \cite{focal_loss} is a variant of CE that is specifically designed to address the class imbalance problem. 
It aims to reduce the relative loss for well-classified examples while emphasizing training on hard, misclassified ones. 
Although Focal loss does not directly consider the class distribution, it automatically distinguishes between hard and easy samples. This proves beneficial in correctly predicting underrepresented classes. Notably, this solution gives more importance to errors (i.e., wrongly predicted tokens) than cross-entropy.

\subsection{Region-based losses}
This family of loss functions optimizes the model weights according to the differences between two mathematical sets.

\paragraph{Dice Loss.}
It is the main representative of the region-based loss family. Dice Loss (DL)~\citep{dice_loss} optimizes the Dice Score (DS) between two sets.
DL directly maximizes a soft version of the Dice Score. It assigns different weights to errors and correct predictions.
However, correct predictions are deemed more relevant than wrong predictions; therefore, errors may not be sufficiently penalized.

\paragraph{Generalized Dice Loss.}
A generalization of the Dice score \citep{gdice_score} and the corresponding Generalized Dice Loss (GDL) was proposed to consider each class's volume. This formulation proposes to self-adjust the weight of each class for each sample to address the class imbalance issue.

\paragraph{Lovász Loss.}
Lovász Loss (LL) \cite{lovasz_loss} is a surrogate loss deriving from the Jaccard Index (or Intersection-over-Union).
LL takes into account both errors and correct predictions. In contrast to Dice loss, which assigns more weight to correctly classified samples, the formulation of Lovász loss allows for an adequate penalty for misclassifications.
In many language tasks, the aim is not only to penalize errors but also to force the system to avoid introducing extra tokens or omitting certain tokens. This objective can be reached by optimizing the Jaccard Index.
We claim that optimizing this objective can be particularly beneficial for the mathematical reasoning task if we ask the model to generate both the final answer and the intermediate reasoning steps. In this case, the intermediate steps must adhere to a stringent structure in terms of syntax (i.e., Math is a formal language) and content (i.e., there are usually not many alternative procedures to get the final answer). This makes the task suitable for optimization using Lovász loss.

\subsection{Compound Losses}
Compound losses are created by combining other loss functions, resulting in a more complex (and possibly more representative) objective function.

\paragraph{Self-Adjusting Dice Loss.}
We also evaluate Self-Adjusting Dice Loss (SADL) \citep{dice_nlp}, which combines the intuitions of Dice and Focal losses. The rationale behind introducing the Focal component in the Dice Loss is to address the imbalance problem between well-classified and misclassified tokens, which is not adequately covered by Dice loss. 

\subsection{Loss application to language generation}

\begin{figure}[htb]
    \centering
    \includegraphics[width=0.75\linewidth]{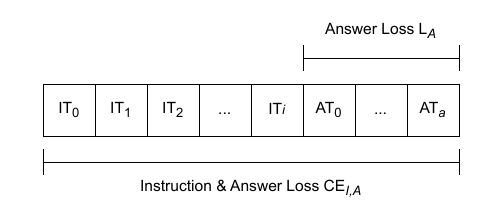}
    \caption{A graphical sketch of how the combined loss is applied to instruction $I$ and answer $A$. \textit{IT}s are instruction tokens, \textit{AT}s are answer tokens.}
    \label{fig:combo-loss}
\end{figure}

Let $I$ and $A$ be the instruction and its corresponding answer. Let $i$ and $a$ be the number of tokens in $I$ and $A$, respectively. 
We define the language modeling loss as a convex combination \citep{combo_loss} of CE and one of the various loss functions L under consideration (i.e., FL, DL, GDL, SADL, and LL):
$\mathcal{L}=\lambda\text{CE}_{I,A}+(1-\lambda)\text{L}_A$, where $\lambda \in [0, 1]$. CE is applied to both the $I$'s and $A$'s tokens, while L is applied only to the $A$'s tokens of the answer (i.e., ground truth), as shown in \Cref{fig:combo-loss}. This approach emphasizes the actual target sequence of interest, which follows a more rigid structure.  Applying the second component to the instruction tokens may wrongly emphasize underrepresented tokens that are not useful in this case.

\subsection{Evaluation Metrics}
We employ both standard metrics that consider the final result only (i.e., Exact Match (EM)) and metrics that are specifically tailored to assess the reasoning steps. Since reasoning step metrics are suited to MWP only, they will be reported only for GSM8K and MathQA datasets.

\paragraph{Metrics for the reasoning steps.}
We adopt the ROSCOE metrics~\cite{roscoe} and general purpose metrics to evaluate the correctness of intermediate reasoning steps, given the systematic and precise nature of mathematical language:
Jaccard Index (or IoU, in short) (see \Cref{sec:loss_formulation}), and Commutative IoU (C-IoU), which we define as a variant of IoU that accounts for the commutative property of mathematical operations. These metrics are calculated between predicted rationales and ground truth reasoning steps. Unlike ROSCOE, adopting this approach eliminates reliance on external models, thus circumventing potential limitations inherent to the models used.

ROSCOE metrics consider four perspectives: Semantic Alignment (SA), Semantic Similarity (SS), Logical Inference (LI), and Language Coherence (LC). Each metric ranges between zero (worst) and one (best). While, for completeness, we evaluate all the proposed metrics, we argue that LC metrics may not be suitable for assessing mathematical steps, as they are not expressed in natural language.

\section{Experimental Results}
\label{sec:exp}
We perform an extensive experimental evaluation on two tasks for a total of four datasets, five models, and five loss functions. In the following, we summarize the main results reporting the average across models.
Detailed results are available in \Cref{sec:complete_results}.

\subsection{Datasets}
We selected four datasets, each including at least training and validation sets, neglecting those containing only the test set (being designed for zero-shot benchmarking). 

We selected two MWP datasets: GSM8K \citep{cobbe2021gsm8k} and MathQA \citep{mathqa}. We have chosen these
datasets %
since they include both the final result and the operational annotations (i.e., reasoning steps) leading to the final answer. 
We also selected two multiple-choice datasets included in the HELM benchmark: OpenBookQA \citep{openbookqa} and HellaSwag \citep{hellaswag}. We consider these QA datasets since answers are mainly based on reading comprehension rather than relying on prior knowledge of the LLM. 

Detailed information on the considered datasets, including their training/validation/test set splits, are available in~\Cref{sec:dataset_statistics}.

\subsection{Competitors}

\begin{table*}[]
    \centering
    \resizebox{\linewidth}{!}
    {
        \begin{tabular}{lllcccc}
        \toprule
            \textbf{Model} & \textbf{Training Strategy} & \textbf{\# Training Samples} & \textbf{\# Params} & \textbf{\# Updated Params} & \textbf{Generalist} & \textbf{Domain} \\ 
            \midrule
            Ours & IT & $\approx$ 5K-40K & 3B-7B & < 1\% & \checkmark & -- \\ 
            MAmmoTH~\cite{mammoth} & IT & $\approx$ 260K & 7B & 100\% & \xmark & Math\\ 
            WizardMath~\cite{wizardmath} & RLEIF & $\approx$ 96K & 7B & 100\% & \xmark & Math\\ 
            WizardLM \cite{wizardlm} & EI & $\approx$ 286K & 7B & 100\% & \checkmark & -- \\ 
            Llemma~\cite{llemma} & P & $\approx$ 30M & 7B & 100\% & \xmark & Math \\ 
            MetaMath~\cite{metamath} & IT & $\approx$ 395K & 7B & 100\% & \xmark & Math \\ 
            Mistral v0.2~\cite{mistral} & IT & ? & 7B & 100\% & \checkmark & -- \\
            GPT4~\cite{OpenAIGPT42023} & RLHF & ? & ? & ?  & \checkmark & -- \\ 
            \bottomrule
        \end{tabular}
    }
    \caption{Competitors details. \textit{IT} stands for Instruction Tuning, \textit{P} for pre-training, \textit{RLHF} for Reinforcement Learning from Human Feedback, \textit{EI} for Evol-Instruct, and \textit{RLEIF} for Reinforcement Learning from Evol-Instruct Feedback.}
    \label{tab:competitors_details}
\end{table*}

In our selection of competitors, %
we considered two key criteria: models with comparable sizes in terms of billions of parameters and open-source solutions rather than closed-source alternatives. 
As shown in \Cref{tab:competitors_details}, we selected open-source solutions trained with different strategies: MAmmoTH~\cite{mammoth}, WizardMath~\cite{wizardmath}, WizardLM \cite{wizardlm}, Llemma~\cite{llemma}, MetaMath \cite{metamath}, and Mistral \cite{mistral}.
For the sake of completeness, we also included a closed-source state-of-the-art model (GPT-4 \cite{OpenAIGPT42023}).
Some models employ domain-specific training, while other approaches are more generalist.

These models employ diverse training strategies, including continual pre-training, instruct-following fine-tuning, Reinforcement Learning From Evol-Instruct Feedback, Evol-Instruct, and Reinforcement Learning from Human Feedback. Consequently, these approaches often necessitate larger datasets or human interventions or rely on other language models for training.

The training data sizes vary significantly, ranging from a hundred thousand samples to millions of samples for models like Llemma. While most competitors updated 100\% of their parameters during training, our approach involves updating less than 1\% of the 3-7 billion parameters using LoRA.

\subsection{Models}
We employ the following LLMs with a number of parameters ranging from 3B to 7B:
RedPajama-Incite-3B \citep{redpajama}, StableLM-3B \citep{stablelm}, RedPajama-Incite-7B \citep{redpajama}, Falcon-7B \citep{falcon}, and Llama-2-7B \citep{touvron2023llama}. Except for Llama-2 (which is selected as one of the most well-known open-source models), the other ones are selected with the following criteria: (1) They are open-source; (2) They 
show promising results according to HELM benchmark~\citep{helm};
(3) The majority of their training datasets are public or clearly stated to avoid overlapping with the analyzed datasets;
(4) We consider only the pre-trained version (without any instruction tuning or tuning by human preferences).

More details about the selected models can be found in \Cref{sec:model_summary}.

\subsection{Experimental settings}
We set the number of training steps to around $25000$ and the batch size to 2. 
We employ Low-Rank Adaptation~\cite{lora}, AdamW optimizer~\cite{DBLP:journals/corr/abs-1711-05101}, and a linear learning rate scheduler with a warmup of 500 steps. %
Further information about the experimental settings and implementation details are given in \Cref{sec:implementation_details}.

\begin{table}[htb]
    \centering
    \subfloat[Exact Match\label{tab:em_results}]{\resizebox{\linewidth}{!}{%
\begin{tabular}{@{}l|cccc|c@{}}
\toprule
\textbf{Loss} & \textbf{HellaSwag} $\uparrow$  & \textbf{OpenBookQA} $\uparrow$ & \textbf{GSM8K} $\uparrow$ &    \textbf{MathQA}  $\uparrow$ & \textbf{MR}  $\downarrow$         \\ \midrule
CE   & 47.36 & 75.60         & 15.83 &   5.12     &        3.33    \\
FL   & \textbf{71.68} & 80.88   & 15.41  &  \textbf{5.52}   & 2.43                 \\
GDL  & 47.39 & 75.08              & 15.00 & 5.04   &  3.93     \\ 
LL   & 58.08 & \textbf{82.80}       & \textbf{17.76} & 4.76 & \textbf{1.90}           \\
SADL & 41.83 & 67.40                 & 15.91&  4.48   &  3.42 \\ \bottomrule
\end{tabular}}} \hfill
    \subfloat[Reasoning Step metrics \label{tab:rs_results}]{\resizebox{\linewidth}{!}{%
\begin{tabular}{@{}l|l|cc|cccc@{}} \toprule
	                                                            &                       & \multicolumn{2}{c|}{\textbf{General purpose}} & \multicolumn{4}{c}{\textbf{ROSCOE}} \\
	         & \textbf{Loss}                  & \textbf{IoU} $\uparrow$                                & \textbf{C-IoU} $\uparrow$                   & \textbf{SA} $\uparrow$    & \textbf{SS}  $\uparrow$  & \textbf{LI} $\uparrow$   & \textbf{LC}  $\uparrow$  \\
	\midrule\multirow{5}{*}{\rotatebox[origin=rb]{90}{GSM8K}}                                                     & CE                    & 15.52                                        & 19.27                             & 81.14          & 65.75          & 34.91          & 37.58          \\
	                                                 & FL   & 15.09                                        & 18.71                             & 81.38          & \textbf{66.67} & \textbf{36.74} & 37.60 \\
	                                                 & GDL  & 15.15                                        & 18.70                             & 81.08          & 65.73          & 34.70          & 37.60 \\
	                                                 & LL   & \textbf{17.39}                               & \textbf{21.10}                    & \textbf{81.39} & 66.33          & 36.00          & 37.46          \\
	                                                & SADL & 15.64                                        & 19.51                             & 81.33          & 66.29          & 35.47          & \textbf{37.62}          \\
	\midrule \multirow{5}{*}{\rotatebox[origin=rb]{90}{MathQA}} & CE                    & 36.72                                        & 36.78                             & 85.12          & 68.43          & 24.21          & 38.86          \\
	                                                & FL   & 33.73                                        & 33.79                             & 85.29          & 68.39          & 23.75          & 38.80          \\
	                                                 & GDL  & 36.30                                        & 36.36                             & 85.07          & 67.05          & 21.01          & 38.90          \\
	                                                & LL   & \textbf{43.25}                               & \textbf{43.31}                    & \textbf{85.76} & \textbf{70.03} & \textbf{28.68} & 38.75          \\
	                                                & SADL & 34.18                                        & 34.23                             & 84.97          & 67.05          & 20.42          & \textbf{38.95} \\
	\bottomrule
\end{tabular}}} \hfill
    \caption{Macro-average achieved on analyzed datasets.}%
    \label{tab:math_aggregate}
\end{table}

\subsection{Answer generation results}
Considering the exact match, as shown in \Cref{tab:em_results}, the CE-only setting is a suboptimal choice in every case.
Based on the mean rank across all models and datasets (i.e., the average rank of each loss), the best losses for these tasks are the Focal and Lovász losses. They show a difference of 0.9 and 1.43, respectively, compared to the CE rank.

The effectiveness of the Focal and Lovász losses is likely due to their distinct approaches to handling prediction errors. The Focal loss underestimates the loss contributions of well-predicted samples based on class distribution, while the Lovász loss penalizes wrong predictions without suppressing well-predicted samples according to their distributional behavior.

\subsection{Reasoning steps generation results}
On both MWP datasets, considering reasoning metrics, the combination with Lovász loss consistently outperforms the CE-only setting as shown in \Cref{tab:rs_results}. 
Also in this case, it achieves the best performance, likely thanks to the effect of misclassified sample penalties.   
Specifically, while cross-entropy and Focal loss aim to maximize global accuracy, LL aims to maximize the global IoU, i.e., it considers both the absence of extra tokens and the presence of missing tokens.

The results on MathQA and GSM8K show that the final answer tends to be wrong in many cases (low EM values), while the reasoning steps tend to be quite accurate (high or medium-high reasoning step metrics). 
This highlights that the models generally struggle to correctly predict the final result despite showing a good capability in formulating the mathematical reasoning steps. 

The complete set of results on all datasets for all models and metrics are available in \Cref{sec:complete_results}, along with statistical tests for significance between cross-entropy and the other loss functions.

\paragraph{Correlation analysis between reasoning step metrics.}
We investigate the Pearson's correlation between the ROSCOE metrics, EM, and IoU to understand if the optimization of this last metric goes in the same direction as more complex ones. 
As expected, IoU is positively correlated (values in range $[0.5, 0.7]$) with many Semantic Alignment metrics, as Reasoning Alignment, External Hallucination, Redundancy, Common Sense Error, Missing Step, and with a Sematic Similarity metric, i.e., Semantic Coverage Chain.
This confirms that optimizing IoU (through the Lovász loss) is a reasonable proxy to optimize the reasoning step metrics. More details are given in \Cref{sec:correlation_roscoe}.

\paragraph{Error type analysis in MWP.}
\label{sec:mwp_errors}
We analyze the most common mistakes observed in MWP reasoning steps. 
We consider the following metrics covering complementary types of reasoning errors\footnote{To the best of our knowledge, there are no standard metrics to evaluate mathematical reasoning.}:

\begin{itemize}[noitemsep]
    \item \text{Extra Step} (ES): proportion of predicted rationales not included in the gold annotations: $\text{ES} = |PS - GTS|/|PS|$
    \item \text{Missing Step} (MS): proportion of gold rationales not generated by the model:\\
    $\text{MS} = |GTS - PS|/|GTS|$
    \item \text{Wrong Operators} (WO): proportion of predicted rationales with correct operands but wrong sign according to the gold rationales: $\text{WO} = |PS_{wo}|/|E|$
    \item \text{Inverted Operands} (IO): proportion of predicted rationales in which the operands have an incorrect position, considering non-commutative operations: $\text{IO} = |PS_{io}| / |E|$
\end{itemize}

\noindent where $GTS$ and $PS$ are the ground truth and predicted reasoning steps, $PS_{wo}$ and $PS_{io}$ are predicted steps with a wrong operator and inverted operands, and $E$ is the set of errors, i.e., the set of predicted reasoning steps that do not match the gold rationales.

The results are summarized in Table~\ref{tab:errors}. Lovász loss yields the lowest percentages of errors across most error types, particularly in reducing the amount of missing steps. The errors related to wrong operators and inverted operands affect approximately only 4-5\% of the reasoning steps for all loss functions. Overall, generating fully accurate reasoning chains remains challenging, but losses such as Lovász loss can help mitigate certain types of errors, making it a preferable training loss to cross-entropy.

\begin{table}[htb]
    \centering
    \resizebox{0.8\linewidth}{!}{%
\begin{tabular}{@{}l|cccc@{}}
\toprule
\textbf{Loss} & \textbf{ES} $\downarrow$ & \textbf{MS} $\downarrow$ & \textbf{WO} $\downarrow$ & \textbf{IO} $\downarrow$ \\ \midrule
CE   & 67.60\%     & 67.78\%       & 4.68\%          & 5.13\%            \\
FL   & 67.85\%    & 68.48\%       & \textbf{4.22\%}          & \textbf{4.66\%}            \\
GDL  & 68.30\%    & 66.95\%       & 4.57\%          & 5.00\%            \\
LL   & \textbf{62.87\%}     & \textbf{62.83\%}       & 4.27\%          & \textbf{4.66\%}            \\
SADL & 70.40\%    & 67.32\%       & 4.71\%          & 5.21\%            \\ \bottomrule
\end{tabular}}
    \caption{Mean errors in mathematical reasoning over all models and datasets.}
    \label{tab:errors}
\end{table}

\subsection{Results on a reduced number of samples}
\label{sec:limited_data}
We evaluate the effectiveness of the proposed approach on each task and dataset using our best model (i.e., StableLM) by reducing the number of training samples to $40\%$ and $10\%$, while also reducing the training duration by the same amount. In \Cref{tab:reduced_data}, we present the average results on each dataset by loss. We show that cross-entropy does not generally yield satisfactory results when the amount of data is reduced. Conversely, losses such as Focal and Lovász demonstrate better capability in extracting desired knowledge even from fewer samples. The trend is the same for both exact match and reasoning step metrics.

\begin{table}[htb]
    \centering
    \subfloat[Exact Match]{\resizebox{0.8\linewidth}{!}{%
\begin{tabular}{@{}l|l|ccccc@{}}
\toprule
                      & \textbf{Dataset}  & \textbf{CE}     & \textbf{GDL}     & \textbf{FL}     & \textbf{LL}     & \textbf{SADL}   \\ \midrule
\multirow{4}{*}{\rotatebox{90}{10\%}} 
                    & HellaSwag & 71.90 & 70.71 & \textbf{79.80} & 79.17 & 77.77 \\
                    & OpenBookQA & \textbf{80.40} & 75.00 & 79.80 & 75.80 & 74.00 \\
                    & GSM8K    & 13.72 & 13.57 & 13.57 & \textbf{15.31} & 13.27 \\
                    & MathQA   & 5.61 & 5.54 & \textbf{6.59} & 6.03 & 5.95 \\
                        \midrule
\multirow{4}{*}{\rotatebox{90}{40\%}} 
                    & HellaSwag & 81.72 & 82.05 & \textbf{90.82} & 86.78 & 84.07 \\
                    & OpenBookQA & 83.00 & 83.00 & \textbf{84.60} & 83.00 & 82.20 \\
                    & GSM8K    & 21.68 & 21.76  & 23.88 & \textbf{26.08} & 20.77 \\
                    & MathQA   & 7.08 & 5.12 & \textbf{8.29} & 7.80 & 3.47 \\
                      \bottomrule
\end{tabular}}} \\
    \subfloat[Intersection-over-Union]{\resizebox{0.7\linewidth}{!}{%
\begin{tabular}{@{}l|l|ccccc@{}}
\toprule
                      & \textbf{Dataset}  & \textbf{CE}     & \textbf{GDL}     & \textbf{FL}     & \textbf{LL}     & \textbf{SADL}   \\ \midrule
\multirow{2}{*}{\rotatebox{90}{10\%}} %
                      & GSM8K   & 13.63  & 13.55  & 13.33 & \textbf{14.59} & 13.24 \\
                      & MathQA   & 9.37  & 8.85  & 10.54 & \textbf{10.66} & 8.85 \\
                        \midrule
\multirow{2}{*}{\rotatebox{90}{40\%}} %
                      & GSM8K   & 18.98  & 18.72  & 19.73 & \textbf{21.86} & 18.11 \\
                      & MathQA  & 35.70  & 36.88  & 37.04 & \textbf{40.01} & 34.61 \\
                      \bottomrule
\end{tabular}}}
    \caption{Results of the best-performing model on different training dataset subsets (10\% and 40\%).} 
    \label{tab:reduced_data}
\end{table}

\subsection{Comparison between CE-Only and Loss-By-Task Instruction Tuning}

To evaluate the effectiveness of our approach in an instruction-tuning scenario (similar to \cite{mammoth}), we train our model on a combined dataset containing task-specific samples from all previously mentioned datasets. We compare the results achieved 
by cross-entropy alone and combined with the other loss functions considered. 
According to previous sections, we selected Lovász for MWP and Focal for QA. 
The results in \Cref{tab:it_mode} confirm our strategy is still effective in a dataset composed of different tasks.

\begin{table}[htb]
    \centering
    \subfloat[Exact Match]{\resizebox{\linewidth}{!}{\begin{tabular}{c|cccc}
     \toprule
     \textbf{Loss} & \textbf{HellaSwag} & \textbf{OpenBookQA} & \textbf{GSM8K} & \textbf{MathQA} \\ \midrule
     CE   & 37.69 & 41.08 & 10.06 & 3.28 \\
     Loss-By-Task & \textbf{66.92} & \textbf{49.31} & \textbf{11.77} & \textbf{3.84} \\ \bottomrule
\end{tabular}}} \\
    \subfloat[ROSCOE metrics]{\resizebox{\linewidth}{!}{\begin{tabular}{c|cccc|cccc}
     \toprule
      & \multicolumn{4}{|c|}{\textbf{GSM8K}} & \multicolumn{4}{|c}{\textbf{MathQA}} \\
     \textbf{Loss} & \textbf{SS} $\uparrow$ & \textbf{SA} $\uparrow$ & \textbf{LI} $\uparrow$ & \textbf{LC} $\uparrow$ & \textbf{SS} $\uparrow$ & \textbf{SA} $\uparrow$ & \textbf{LI} $\uparrow$ & \textbf{LC} $\uparrow$ \\\midrule
     CE   & 66.95 & 81.13 & 38.25 & \textbf{37.80} & 72.02 & 84.76 & 30.26 & 39.25 \\
     Loss-By-Task & \textbf{67.02} & \textbf{81.23} & \textbf{38.94} & 37.66 & \textbf{72.84} &\textbf{ 84.80} & \textbf{32.09} & \textbf{40.16} \\ \bottomrule
\end{tabular}}}
    \caption{Mean performance over all datasets in Instruction Tuning mode.}
    \label{tab:it_mode}
\end{table}

\subsection{Comparison with the state of the art}

\begin{table}[htb]
    \centering
    \subfloat[Exact Match]{\resizebox{\linewidth}{!}{\begin{tabular}{@{}l|cccc|c@{}}
\toprule
\textbf{Model} & \textbf{GSM8K} $\uparrow$ & \textbf{MathQA} $\uparrow$ & \textbf{HellaSwag} $\uparrow$ & \textbf{OpenBookQA} $\uparrow$ & \textbf{MR} $\downarrow$ \\ \midrule
Our Best    & 28.66    & 10.06       & \textbf{85.69}          & \textbf{87.20}   &  \underline{2.33}      \\
MAmmoTH  & \underline{37.76}    &   \underline{15.51}      &   7.30        &    3.60   & 4.00    \\
WizardMath     &  \underline{46.10}   &  \textbf{32.43}      &   \underline{36.84}        & \underline{60.00} &  \textbf{2.25}         \\
WizardLM & 9.02 & 3.84 & 26.81 & 33.40 & 5.50 \\
Mistral & 19.64 & 9.76 & \underline{49.26 }& \underline{74.40} & \underline{3.50} \\ 
Llemma \ding{199}  & 30.33 & 9.53 & 24.47 & 21.20 & 5.00 \\
MetaMath & \textbf{60.27} & \underline{14.43} & 14.56 & 19.20 & 4.25 \\
\midrule
GPT4 \ding{93} & 93.20 & -- & 95.30 & 96.00 & -- \\\bottomrule
\end{tabular}}} \\
    \subfloat[ROSCOE metrics]{\resizebox{\linewidth}{!}{\begin{tabular}{@{}l|cccc|cccc|c@{}}
\toprule
      & \multicolumn{4}{|c|}{\textbf{GSM8K}} & \multicolumn{4}{|c}{\textbf{MathQA}} & \\ 
\textbf{Model} & \textbf{SS} $\uparrow$ & \textbf{SA} $\uparrow$ & \textbf{LI}$\uparrow$ & \textbf{LC}  $\uparrow$& \textbf{SS} $\uparrow$& \textbf{SA} $\uparrow$ & \textbf{LI} $\uparrow$ & \textbf{LC} $\uparrow$ & \textbf{MR} $\downarrow$\\ \midrule
Our Best    & 66.10 & \underline{81.76} & \underline{35.70} & 24.10  & \textbf{67.03} & \textbf{86.14} & \underline{24.30} & 25.93 & \underline{3.38}          \\
MAmmoTH   & \underline{66.46} & 81.02 & 10.29 & 24.48  & \underline{64.70} & \underline{80.02} & 17.45 & 23.45 & 4.88       \\
WizardMath  & 64.18 & 80.18 & 15.18 & \underline{27.36}  & 63.36 & 79.71 & 5.60 & \underline{27.88} &  4.00   \\
WizardLM  & 63.71 & 80.05 & 11.44 & \textbf{27.45} & 64.25 & 79.94 & 14.63 & \underline{27.48} & 4.75\\
Mistral  & 63.56 & 81.13 & 13.87 & \underline{26.26}  & 62.97 & \underline{80.52} & 10.58 & 26.76 &  4.75 \\ 
Llemma \ding{199}   & \textbf{74.50} & \textbf{85.70} & \textbf{46.74} & 25.36  & 61.96 & 79.32 & \textbf{66.59} & \textbf{36.22} & \textbf{3.00} \\
MetaMath  & \underline{66.71} & \underline{82.50} & \underline{35.53} & 26.04  & \underline{64.80} & 80.01 & \underline{20.50} & 26.26 & \underline{3.25}\\
\midrule
GPT4 \ding{93} & -- & -- & -- & --  & -- & -- & -- & -- & -- \\\bottomrule
\end{tabular}

}}
    \caption{Competitors results on analyzed datasets.\\\ding{93} indicates results taken from other papers \cite{helm,OpenAIGPT42023} and \ding{199} indicates model tested in 8-shots. The \textbf{best}, \underline{second-best}, and \underline{third-best} results are indicated in each column.}
    \label{tab:comparison_table}
\end{table}

As shown in Table~\ref{tab:comparison_table}, our proposed model achieves the best results in 2 out of 4 datasets. In contrast, domain-specific models, such as MAmmoTH and Llemma, experience a notable degradation in performance when evaluated on closed-ended QA datasets. Our proposed approach achieves comparable performance to WizardMath according to the mean rank (MR), proving its effectiveness across various scenarios without employing any additional steps after fine-tuning (e.g., tuning by preferences). Regarding rationale generation, our best model ranks in the top 3 positions according to the mean rank. 
Although WizardMath and Mistral are the best-performing in terms of exact match, they exhibit the lowest performance according to the ROSCOE metrics.
This confirms the fact that providing the right answer does not necessarily imply the correct reasoning.

\section{Conclusion and Future Work}

In our work, we applied semantic segmentation losses to improve the fine-tuning of LLMs for mathematical reasoning and closed-ended question-answering. Our results show that using appropriate loss functions during fine-tuning can boost performance without extra data or human feedback. In practice, this suggest a promising pathway for more efficient and accessible training processes. Future work will focus on designing new task-specific loss functions and exploring other tasks.

\section*{Limitations}
We analyzed only English datasets from the mathematical reasoning and reading comprehension domains. Additional experiments on other languages and tasks would strengthen the generalizability of our findings. 
It is worth noting that we limited our analysis to existing loss functions in computer vision, which may be suboptimal for the tasks under consideration. We focused on tasks with strong constraints to verify the effectiveness of the analyzed loss functions; however, this approach may pose limitations in datasets with more open-ended solutions lacking well-defined patterns.

Our model choice was based on the available resources, and we tested only 3B and 7B models. Although we could expect similar findings with larger models, we cannot confirm this claim.

\section*{Ethics Statement}
From our understanding, the datasets employed in this study do not contain any personal information, but they can contain some harmful or inappropriate content. This claim can be extended to the employed models, which could provide non-factual, biased, harmful, or inappropriate answers. Their usage is subject to the limitations stated in their respective technical reports and licenses. The generated answers are not intended to offend or harm anyone. Language models have environmental impacts due to the high computing requirements during pre-training and fine-tuning. We have made efforts to be computationally responsible by reusing open-sourced pre-trained models and employing efficient fine-tuning methods such as LoRA~\cite{lora}. The gains from improved losses help amortize the resource costs over higher utility. Overall, we have made reasonable efforts to ensure the transparency and auditability of our experimental methodology.

\bibliography{bibliography}

\newpage
\appendix
\section*{Appendices}
In this supplementary material, we provide additional details as follows:
\begin{itemize}
    \item \Cref{sec:loss_formulation}: Loss Function Formulations
    \item \Cref{sec:dataset_statistics}: Dataset Statistics
    \item \Cref{sec:token_distribution}: Token Distribution
    \item \Cref{sec:model_summary}: Model Summary
    \item \Cref{sec:complete_results}: Extended Results
    \item \Cref{sec:correlation_roscoe}: Correlation between General purpose Metrics and ROSCOE Metrics
    \item \Cref{sec:implementation_details}: Implementation Details
    \item \Cref{sec:qualitative_input}: Prompt Examples
\end{itemize}

\section{Loss Function Formulations}
\label{sec:loss_formulation}
For the sake of simplicity, 
hereinafter we will consider the binary formulation. 
However, the loss formulations can be straightforwardly extended to the multi-class scenario. 

\paragraph{Cross Entropy Loss}
Accuracy (AC) and Cross-Entropy Loss (CE) are defined as follows:
\begin{align}
    \text{AC} = \frac{1}{N}\sum_i^N 1(\hat{y}_i = y_i) \\
    \text{CE}(p_t) = - \log (p_t) 
\end{align}
\noindent where $N$ is the total number of samples, $\hat{y}_i$ and $y_i$ are the predicted and ground truth class for sample $i$, respectively, and $p_t$ is the probability of the sample belonging to the positive class.

\paragraph{Focal Loss}
Focal Loss (FL) \cite{focal_loss} can be defined as follows:

\begin{equation}
        \text{FL}(p_t) = - (1-p_t)^{\gamma}\log (p_t) \label{eq:focal}
\end{equation}

\noindent where $p_t$ is the probability of the sample belonging to the positive class while $\gamma$ is the Focal suppression parameter.

\paragraph{Dice Loss}
Dice Score and Dice Loss (DL) \cite{dice_loss} are defined as follows: 

\begin{align}
    \text{DS} = \frac{2|\hat Y \cap Y|}{|\hat Y| + |Y|} = \frac{2 TP}{2 TP + FP + FN} \label{eq:dice_score} \\ 
    \text{DL} = 1 - \frac{2 \sum_i p_i y_i}{\sum_i p_i^2 + \sum_i y_i^2} \label{eq:dice_loss}
\end{align}

\noindent where $\hat Y$ and $Y$ are the prediction and ground truth sets, $TP$, $FP$, $FN$ are the numbers of true positives, false positives, and false negatives, respectively,  
$p_i$ is the probability of the sample belonging to the positive class, and $y_i$ is the ground truth label.

\paragraph{Self-Adjusting Dice Loss}
Self-Adjusting Dice Loss (SADL) \cite{dice_nlp} can be expressed as follows:
\begin{equation}
    \text{SADL} =  1 - \frac{2\sum_i(1 - p_i)p_i y_i}{\sum_i(1 - p_i)p_i + y_i}
\end{equation}
\noindent where %
$(1 - p_i)$ is the Focal component in \Cref{eq:focal}.

\paragraph{Generalized Dice Loss}
Generalized Dice Loss (GDL) \cite{gdice_loss} can be expressed as follows:

\begin{equation}
    \text{GDL} = 1 - \frac{2 \sum_l w_l \sum_i p_{il} y_{il}}{\sum_l w_l \sum_i p_{il} + y_{il}}
\end{equation}
where $w_l = 1 / (\sum_i y_{il})^2$ for each class, while $p_i$ and $y_i$ have the same meanings as defined in \Cref{eq:dice_loss}.

\paragraph{Lovász Loss}
Let $\hat Y$ and $Y$ represent the prediction and ground truth sets, respectively. The Jaccard Index (or Intersection-over-Union, IoU) is defined as follows:

\begin{equation}
    \text{IoU} = \frac{|\hat Y\cap Y|}{|\hat Y\cup Y|} = \frac{TP}{TP + FP + FN} \label{eq:jaccard_index}
\end{equation}

Lovász surrogate Loss (LL) \cite{lovasz_loss} has the following expression:

\begin{align}
    \Delta_{J_1} = 1 - \frac{|\{\hat Y = 1\} \cap \{Y = 1\}|}{|\{\hat Y = 1\} \cup \{Y = 1\}|} \\
    HL_i(x_i, y_i) = \max (0, 1 - x_i y_i) \\
    \text{LL} = \overline{\Delta_{J_1}} HL(X, Y)
\end{align}

\noindent where $\Delta_{J_1}$ is the Jaccard loss, $HL$ is the hinge loss, $x_i \in X$ is the prediction logit associated to sample $i$, $y_i \in Y$ with $y_i \in \{-1, 1\}$, and $\overline{\Delta_{J_1}}$ is the Lovász extension of the Jaccard loss.

\section{Dataset Statistics}
\label{sec:dataset_statistics}
\begin{itemize}
    \item GSM8K\footnote{\url{https://huggingface.co/datasets/gsm8k}} \cite{cobbe2021gsm8k} is a dataset of 8.5K high-quality linguistically diverse grade school Math Word Problems. The dataset was created to support answering questions on basic mathematical problems requiring multi-step reasoning. It has 7470 samples in the training set and 1320 in the test set. It is released under the MIT license.
    \item MathQA\footnote{\url{https://huggingface.co/datasets/math_qa}} \cite{mathqa} is a large-scale dataset of Math Word Problems enhancing the AQuA dataset \cite{aqua} by providing fully-specified operational programs for each problem. It comprises 29800, 4480, and 2990 samples in the training, validation, and test sets, respectively. It is released under the Apache-2.0 license.
    \item OpenBookQA\footnote{\url{https://huggingface.co/datasets/openbookqa}} \cite{openbookqa} contains questions that require multi-step reasoning, use of additional common and commonsense knowledge, and rich text comprehension. OpenBookQA is modeled after open-book exams for assessing human understanding of a subject. The training, validation, and test sets contain 4960, 500, and 500 samples, respectively. It is released under the Apache-2.0 license.
    \item HellaSwag\footnote{\url{https://huggingface.co/datasets/Rowan/hellaswag}} \cite{hellaswag} introduced a task of commonsense natural language inference, which consists in selecting the most appropriate conclusion for a sentence from a set of possibilities. It contains 39900 samples in the training set and 10000 in the validation set, which is employed as the test set since the actual test set does not have ground truth. It is released under the MIT license.
\end{itemize}

\section{Token Distribution}
\label{sec:token_distribution}
\begin{figure}[htb]
    \centering
    \includegraphics[width=\linewidth]{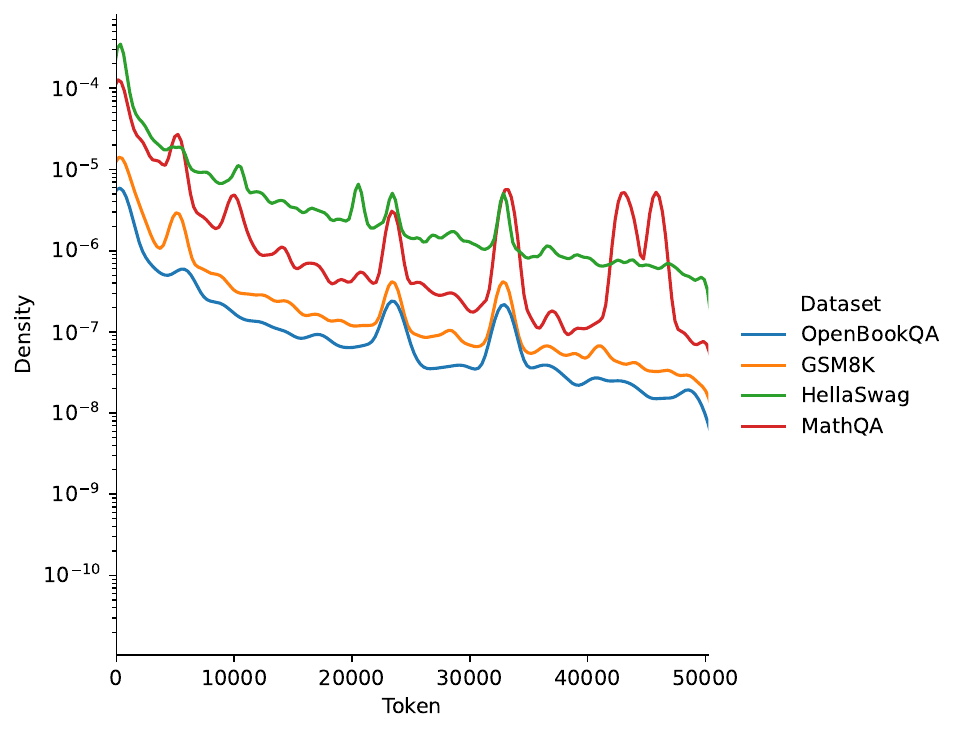}
    \caption{Kernel Density Estimation in log scale for token distributions in GSM8K, MathQA, OpenBookQA, and HellaSwag datasets.}
    \label{fig:token_distribution}
\end{figure}

We report in \Cref{fig:token_distribution} the distribution of tokens across the datasets, highlighting the strong imbalance in tokens. Before the analysis, we excluded all special tokens (25) from the tokenizer.
We plot the density against the token identifier in the log scale to better highlight peaks and differences.

\section{Model Summary}
\label{sec:model_summary}
\Cref{tab:models} summarizes the characteristics of the models used in this work: RedPajama-Incite-3B\footnote{\url{https://huggingface.co/togethercomputer/RedPajama-INCITE-Base-3B-v1}}, StableLM-3B\footnote{\url{https://huggingface.co/stabilityai/stablelm-3b-4e1t}}, RedPajama-Incite-7B\footnote{\url{https://huggingface.co/togethercomputer/RedPajama-INCITE-7B-Base}}, Falcon-7B\footnote{\url{https://huggingface.co/tiiuae/falcon-7b}}, and Llama-2-7B\footnote{\url{https://huggingface.co/meta-llama/Llama-2-7b-hf}}.     
For each of them, the following characteristics are reported: model name, number of parameters, license, availability of the pre-training datasets, and mean win rate according to HELM benchmark \cite{helm}.

\subsection{Competitors}
The competitors chosen are: MAmmoTH\footnote{\url{https://huggingface.co/TIGER-Lab/MAmmoTH-7B}}, WizardMath\footnote{\url{https://huggingface.co/TheBloke/WizardMath-7B-V1.1-GPTQ}}, WizardLM\footnote{\url{https://huggingface.co/TheBloke/wizardLM-7B-HF}}, Llemma\footnote{\url{https://huggingface.co/EleutherAI/llemma_7b}}, MetaMath\footnote{\url{https://huggingface.co/meta-math/MetaMath-7B-V1.0}}, Mistral-7B\footnote{\url{https://huggingface.co/mistralai/Mistral-7B-Instruct-v0.2}}, and GPT-4. We employed the settings and prompts suggested by the authors of the original papers.

MAmmoTH is released under the MIT license. Mistral is released under the Apache 2.0 license. The other models are released under the Llama 2 license.

\begin{table*}[htb]
\centering
    \begin{tabular}{@{}lcllc@{}}
    \toprule
    \textbf{Model }            
    & \textbf{\# Parameters} 
    & \textbf{License}       
    & \textbf{Pre-Training Datasets} 
    & \textbf{HELM Win Rate}\\ 
    \midrule
    
    RedPajama-Incite
    & 3B
    &  Apache 2.0   
    & Public            
    & 0.311        \\
     
    StableLM
    & 3B         
    &  CC BY-SA-4.0 
    & Public             
    & --            \\
    RedPajama-Incite
    & 7B         
    &  Apache 2.0   
    & Public             
    & 0.378        \\
    Falcon   
    & 7B         
    &  Apache 2.0   
    & 90\% Public        
    & 0.378        \\ 
    Llama-2       
    & 7B         
    &  Llama-2      
    & Public             
    & 0.607        \\
    \bottomrule
    \end{tabular}
\caption{Model characteristics.}
\label{tab:models}
\end{table*}

\section{Extended Results}
\label{sec:complete_results}
In the following, we report the extended results for the mathematical reasoning and question-answering tasks.

\subsection{Complete results on MWP}
In~\Cref{tab:math_full_results,tab:gsm_full_results}, we present the detailed performance of each model and loss function on MWP datasets. We use McNemar's test for exact match and t-tests \cite{stat_test} for other metrics to determine if differences are statistically significant. According to our metrics in GSM8K, Lovász provides the best mean performance across all models, except on Falcon, where Self-Adjusting Dice yields the best results. However, differences are not statistically significant, likely due to the model's limitations. On MathQA, Lovász achieves the best performance across most metrics, while for exact match, Focal performs best 2 out of 5 times. The results for ROSCOE metrics in \Cref{tab:roscoe} across both MWP datasets show that Lovász performs best in most metrics, as highlighted by the mean rank as well.

\subsection{Complete results on Question Answering}
In~\Cref{tab:qa_results}, we present the detailed performance of each model and loss function on closed-ended QA datasets. We perform McNemar's test \cite{stat_test} to assess whether differences compared to cross-entropy loss alone are statistically significant. In 6 cases, Lovász loss provides the best improvements, while in 4 cases, Focal loss obtains the best results. The main differences are seen when Lovász fails, whereas Focal still gets improvement. In the opposite case, the results are similar.

\section{Correlation between General purpose Metrics and ROSCOE Metrics}
\label{sec:correlation_roscoe}
In \Cref{tab:roscoe_correlation}, we report the Pearson's correlation analysis between Exact Match (EM), Precision (Prec), Recall (Rec), Dice Score (DS), Intersection-over-Union (IoU), Commutative Intersection-over-Union (C-IoU) and ROSCOE metrics, showing medium-high correlation values.
Reasoning Alignment (RA) and Redundancy (RD) exhibit the strongest correlations with the general-purpose metrics. Common Sense Error (CSE) and Semantic Coverage Chain (SCC) demonstrate moderate correlation values. External Hallucination (EH) and Missing Steps (MS) show a moderate correlation as well.

\begin{table}[htb]
    \centering
    \resizebox{\linewidth}{!}{
\begin{tabular}{@{}l|cccccc@{}}
    \toprule
        & \textbf{EM} & \textbf{IoU} & \textbf{Prec} & \textbf{Rec} & \textbf{DS} & \textbf{C-IoU} \\ \midrule
    RA (SA) & 0.1615      & 0.6582       & 0.6891        & 0.6076       & 0.6739      & 0.6698         \\
    EH (SA)  & 0.1425      & 0.6058       & 0.6186        & 0.5115       & 0.5919      & 0.6074         \\
    RD (SA) & 0.1607      & 0.6781       & 0.6911        & 0.5674       & 0.6600      & 0.6828         \\
    CSE (SA) & 0.1559      & 0.5583       & 0.5314        & 0.5741       & 0.5596      & 0.5608         \\
    MS (SA) & 0.1744      & 0.6461       & 0.6138        & 0.6595       & 0.6463      & 0.6523         \\
    SCC  (SS)& 0.1345      & 0.5403       & 0.5501        & 0.5005       & 0.5484      & 0.5495         \\ \bottomrule
    \end{tabular}
    }
    \caption{Pearson's correlation between reasoning metrics (ROSCOE) and standard ones (EM, IoU, Prec, Rec, DS, C-IoU) over all samples.}
    \label{tab:roscoe_correlation}
\end{table}

\section{Implementation Details}
\label{sec:implementation_details}

Based on preliminary experiments, we set the language modeling loss mixing parameter to $\lambda=0.6$. The Focal suppression parameter was set to $\gamma = 2$. The maximum learning rate was set to $1e-4$ for all datasets, except in GSM8K, for which it was set to $1e-5$.

We selected the model checkpoints according to the best validation loss. We train less than $1\%$ of the total model parameters using LoRA. During training, the context size is chosen to include most samples without truncation according to 75\% percentiles: 128 for GSM8K, MathQA, OpenBookQA, and 256 for HellaSwag. We employ gradient accumulation for context size 256.

We employed Transformers and PEFT libraries. Full requirements, versions, and losses' licenses are available in the code repository. For ROSCOE metrics evaluation, we employed the models suggested in the original paper: SimCSE\footnote{\url{https://huggingface.co/facebook/roscoe-512-roberta-base}} for sentence embedding, RoBERTa\footnote{\url{https://huggingface.co/FacebookAI/roberta-base}} as word embedding model, DeBERTa\footnote{\url{https://huggingface.co/MoritzLaurer/DeBERTa-v3-large-mnli-fever-anli-ling-wanli}} as NLI model, RoBERTa\footnote{\url{https://huggingface.co/cointegrated/roberta-large-cola-krishna2020}} as grammar model, and GPT-2\footnote{\url{https://huggingface.co/openai-community/gpt2-large}} as perplexity model.

We ran our experiments on a machine equipped with Intel\textsuperscript{\textregistered} Core\textsuperscript{TM} i9-10980XE CPU, $1$ $\times$ NVIDIA\textsuperscript{\textregistered} RTX A6000 48GB GPU, $128$ GB of RAM running Ubuntu $22.04$ LTS.

\begin{table*}[htb]
    \centering

\begin{tabular}{l|l|cccccc}
\toprule
\textbf{Model}         & \textbf{Loss} & \textbf{EM} & \textbf{IoU}     & \textbf{Prec} & \textbf{Rec}  & \textbf{DS} & \textbf{C-IoU}  \\
\midrule                                                                     
\multirow{5}{*}{RedPajama 3B}  & CE   & 9.33   & 11.03  & 14.66    & 15.51  & 14.69     & 14.76   \\
  & FL   & 9.55   & 11.46  & 15.23    & 16.16  & 15.33     & 15.21   \\
 & GDL  & 9.25   & 11.15  & 14.81    & 15.67  & 14.83     & 14.92   \\
 & LL   & \textbf{11.45$^*$}  & \textbf{12.52$^*$} & \textbf{16.66$^*$}   & \textbf{17.17$^*$} & \textbf{16.52$^*$}    & \textbf{16.53$^*$} \\
 & SADL & 10.16   & 11.76  & 15.80$^*$   & 16.19  & 15.60     & 15.73$^*$  \\ \midrule
\multirow{5}{*}{StableLM 3B}   & CE   & 24.79   & 20.96  & 26.05    & 26.72  & 25.93     & 24.56   \\
  & FL   & 24.79   & 21.81$^*$ & 27.36$^*$   & 27.49  & 26.95$^*$    & 25.51$^*$  \\
   & GDL  & 24.87   & 21.01  & 26.11$^*$   & 26.75  & 25.98     & 24.58   \\
  & LL   & \textbf{28.66$^*$}  & \textbf{24.02$^*$} & \textbf{29.42$^*$}   & \textbf{30.38$^*$} & \textbf{29.38$^*$}   & \textbf{28.15$^*$}  \\
  & SADL & 26.99$^*$  & 21.08  & 26.43    & 27.40  & 26.39     & 25.20   \\ \midrule
\multirow{5}{*}{RedPajama 7B}  & CE   & 16.07   & 15.39  & 19.93    & 20.38  & 19.76     & 19.76   \\
  & FL   & 14.94   & 14.93  & 19.92    & 19.55  & 19.32     & 18.82   \\
  & GDL  & 13.19$^*$  & 13.94$^*$ & 18.27$^*$   & 19.24$^*$ & 18.33$^*$    & 17.94$^*$  \\
  & LL   & \textbf{16.83}   & \textbf{16.66$^*$} & \textbf{21.57$^*$}   & \textbf{21.52}  & \textbf{21.13$^*$}    & \textbf{20.91$^*$}  \\
 & SADL & 13.95$^*$  & 14.94  & 19.32    & 20.41  & 19.44     & 18.85   \\ \midrule
\multirow{5}{*}{Falcon 7B}     & CE   & 4.70   & 11.39  & 14.00    & 20.64  & 16.15     & 14.16   \\
    & FL   & 3.49   & 9.19$^*$ & 11.25$^*$   & 19.47  & 13.69$^*$    & 11.92$^*$  \\
   & GDL  & 4.40   & 11.16  & 13.65    & 20.85  & 15.98     & 13.98   \\
   & LL   & 5.00   & 11.59  & 13.93    & 22.09  & 16.47     & 14.08   \\
  & SADL & \textbf{5.08}   & \textbf{12.04}  & \textbf{14.37}    & \textbf{23.70}  & \textbf{17.18}     & \textbf{15.00}   \\ \midrule
\multirow{5}{*}{Llama-2 7B} & CE   & 24.28   & 18.85  & 23.62    & 23.92  & 23.35     & 23.13   \\
 & FL   & 24.28   & 18.07  & 22.61    & 23.78  & 22.76     & 22.07   \\
 & GDL  & 23.29   & 18.47  & 23.26    & 23.64  & 23.01     & 22.07   \\
 & LL   & \textbf{26.86$^*$}  & \textbf{22.14$^*$} & \textbf{27.09$^*$}   & \textbf{27.74$^*$} & \textbf{26.93$^*$}    & \textbf{25.83$^*$}  \\
 & SADL & 23.37   & 18.36  & 22.98    & 24.03  & 23.01     & 22.78   \\
\bottomrule
\end{tabular}

    \caption{Results on GSM8K dataset. $^*$ indicates values for which $p < 0.05$.}
    \label{tab:gsm_full_results}
\end{table*}

\begin{table*}[htb]
    \centering
\begin{tabular}{l|l|cccccc}
\toprule
\textbf{Model}         & \textbf{Loss} & \textbf{EM} & \textbf{IoU}     & \textbf{Prec} & \textbf{Rec}  & \textbf{DS} & \textbf{C-IoU}  \\
\midrule  
\multirow{5}{*}{RedPajama 3B}  & CE   & \textbf{3.47}   & 30.26  & 34.20    & 35.32  & 34.07     & 30.29  \\
 & FL   & 2.79   & \textbf{33.11$^*$} & \textbf{37.29$^*$}   & 37.87$^*$ & \textbf{36.88$^*$}    & \textbf{33.16$^*$} \\
 & GDL  & 2.45$^*$  & 28.98$^*$ & 32.96$^*$   & 33.96$^*$ & 32.72$^*$    & 29.06$^*$ \\
 & LL   & 2.83   & 32.83$^*$ & 36.48$^*$   & \textbf{38.44$^*$} & 36.69$^*$    & 32.86$^*$ \\
 & SADL & 2.79   & 26.54$^*$ & 30.35$^*$   & 32.55$^*$ & 30.49$^*$    & 26.58$^*$ \\ \midrule
\multirow{5}{*}{StableLM 3B }  & CE   & 8.21   & 61.98  & 64.86    & 67.39  & 65.36     & 62.02  \\
  & FL   & \textbf{10.06$^*$}  & 61.98$^*$ & 65.43$^*$   & 67.47$^*$ & 65.66$^*$    & 62.04$^*$ \\
  & GDL  & 6.86   & 57.13$^*$ & 60.16$^*$   & 63.61$^*$ & 61.03$^*$    & 57.16$^*$ \\
 & LL   & 7.50   & \textbf{65.73$^*$} & \textbf{68.51$^*$}   & \textbf{70.79$^*$} & \textbf{69.06$^*$}    & \textbf{65.80$^*$} \\
  & SADL & 7.16   & 59.79$^*$ & 62.85$^*$   & 65.31$^*$ & 63.33$^*$    & 59.84$^*$ \\ \midrule
\multirow{5}{*}{RedPajama 7B } & CE   & 7.16   & 40.35  & 44.32    & 45.01  & 43.98     & 40.41  \\
 & FL   & \textbf{8.78$^*$}  & 43.12$^*$ & 47.72$^*$   & 48.28$^*$ & 47.16$^*$    & 43.17$^*$ \\
  & GDL  & 7.05   & 41.21$^*$ & 44.87$^*$   & 45.98$^*$ & 44.77$^*$    & 41.27$^*$ \\
 & LL   & 6.82   & \textbf{46.34$^*$} & \textbf{49.87$^*$}   & \textbf{51.27$^*$} & \textbf{49.92$^*$}    & \textbf{46.41$^*$} \\
  & SADL & 6.10   & 32.41$^*$ & 39.17    & 36.75$^*$ & 36.79     & 32.48$^*$ \\ \midrule
\multirow{5}{*}{Falcon 7B}   & CE   & 5.24   & 11.34  & 13.80    & 21.72  & 15.93     & 11.44  \\
   & FL   & 5.84   & 10.93$^*$ & 12.98$^*$   & 24.59$^*$ & 15.77$^*$    & 11.00$^*$ \\
  & GDL  & 5.69   & 11.07$^*$ & 13.21$^*$   & 22.98$^*$ & 15.63$^*$    & 11.14$^*$ \\
 & LL   & 5.35   & \textbf{12.77}  & \textbf{15.00$^*$}   & \textbf{26.07$^*$} & \textbf{17.67$^*$}    & \textbf{12.87}  \\
 & SADL & \textbf{5.99}   & 10.57$^*$ & 12.62$^*$   & 21.50$^*$ & 14.84$^*$    & 10.63$^*$ \\ \midrule
\multirow{5}{*}{Llama-2 7B } & CE   & 1.51   & 39.69  & 44.34    & 45.45  & 43.98     & 39.75  \\
 & FL   & 0.15$^*$  & 19.51$^*$ & 22.29$^*$   & 30.48$^*$ & 24.43$^*$    & 19.60$^*$ \\
 & GDL  & \textbf{3.17$^*$}  & 43.12$^*$ & 45.56    & 57.74$^*$ & 48.87$^*$    & 43.16$^*$ \\
 & LL   & 1.28   & \textbf{58.56$^*$} & \textbf{61.00$^*$}   & \textbf{66.16$^*$} & \textbf{62.28$^*$}    & \textbf{58.62$^*$} \\
 & SADL & 0.38$^*$  & 41.57$^*$ & 43.45    & 58.87$^*$ & 47.77$^*$    & 41.62$^*$ \\
\bottomrule
\end{tabular}

    \caption{Results on MathQA dataset. $^*$ indicates values for which $p < 0.05$.}
    \label{tab:math_full_results}
\end{table*}

\begin{table*}[htb]
    \centering
\begin{tabular}{@{}l|ccccc@{}}
\toprule
                     & CE     & FL     & GDL    & LL     & SADL   \\ \midrule
Faithfulness             & 81.96 & 81.97 & 81.98 & \textbf{82.21} & 81.96 \\
Informativeness Step     & 80.61 & 81.09 & \textbf{81.11} & 80.82 & 81.10  \\
Faithfulness WW          & 91.84 & 92.61 & \textbf{92.78} & 91.55 & 92.77 \\
Informativeness Chain    & 90.63 & 90.40  & 90.50  & \textbf{90.79} & 90.41 \\
Repetition Word          & 12.59 & 13.58 & 9.80  & \textbf{15.67} & 10.91 \\
Repetition Step          & 14.44 & 16.02 & 12.30  & \textbf{17.40}  & 13.30  \\
Reasoning Alignment      & 92.47 & 92.37 & \textbf{92.67} & 92.61 & 92.60  \\
External Hallucination   & 97.59 & 97.60  & 97.57 & \textbf{97.70}  & 97.58 \\
Redundancy               & 88.71 & 88.60 & 88.69 & \textbf{89.06} & 88.62 \\
Common Sense Error       & 97.91 & 97.87 & \textbf{97.96} & \textbf{97.96} & 97.93 \\
Missing Step             & 89.47 & 89.47 & \textbf{89.89}& 89.82 & 89.74 \\
Semantic Coverage Step   & 98.14 & 98.25 & 98.31 & \textbf{98.32} & 98.27 \\
Semantic Coverage Chain  & 96.21 & 96.17 & \textbf{96.36} & 96.35 & 96.30  \\
Discourse Representation & 42.71 & 42.73 & 41.50  & \textbf{45.68} & 40.95 \\
Perplexity Step          & 0.28 & 0.27 & \textbf{0.28} & 0.26 & 0.27 \\
Coherence Step vs Step   & 16.41 & 17.76 & 14.21 & \textbf{19.00}   & 14.94 \\
Perplexity Chain         & 6.08 & 6.42 & 6.74 & 5.49 & \textbf{6.84} \\
Perplexity Step Max      & 0.14 & 0.13 & 0.14 & 0.14 & \textbf{0.15}\\
Grammar Step             & 94.27 & 94.18 & 94.12 & \textbf{94.28} & 94.18 \\
Grammar Step Max         & 90.32 & 90.02 & 89.95 & \textbf{90.34} & 90.00    \\ \midrule
Mean Rank                & 3.20    & 3.45   & 2.80    & \textbf{1.95}   & 3.20   
\end{tabular}
    \caption{Results using ROSCOE metrics aggregated across models and datasets.}
    \label{tab:roscoe}
\end{table*}

\begin{table*}[htb]
    \centering
    \resizebox{0.5\linewidth}{!}{%
\begin{tabular}{@{}l|l|cc@{}}
\toprule
\textbf{Model}                       & \textbf{Loss}    & \textbf{HellaSwag} & \textbf{OpenBookQA} \\ \midrule
\multirow{5}{*}{RedPajama 3B}  & CE      & 25.26     & 66.60       \\
& FL   & \textbf{45.91$^*$}   & \textbf{78.60$^*$}      \\
& GDL   & 25.39     & 63.80       \\
& LL  & 26.05     & 77.20$^*$      \\
& SADL & 25.79$^*$    & 67.00       \\
\midrule
\multirow{5}{*}{StableLM 3B}         & CE      & 79.69     & 84.00       \\
        & FL   & \textbf{85.69$^*$}    & 85.40       \\
         & GDL   & 80.00      & 82.80       \\
      & LL  & 82.97$^*$    & \textbf{87.20$^*$}      \\
    & SADL & 80.49$^*$    & 82.40       \\ \midrule
\multirow{5}{*}{RedPajama 7B}  & CE      & 25.16     & 74.80       \\
  & FL   & \textbf{73.29$^*$}    & 81.60$^*$      \\
  & GDL   & 25.04     & 75.80       \\
 & LL  & 25.08     & \textbf{83.80$^*$}      \\
 & SADL & 25.10      & 76.60       \\
\midrule
\multirow{5}{*}{Falcon 7B}            & CE      & 24.59     & 69.20      \\
                & FL   & 68.51$^*$    & 77.20$^*$      \\
              & GDL   & 24.94     & 69.20       \\
                 & LL  & \textbf{70.72$^*$}   & \textbf{79.00$^*$}      \\
              & SADL & 26.67$^*$    & 55.00$^*$      \\
\midrule
\multirow{5}{*}{Llama-2 7B}              & CE      & 82.12     & 83.40       \\
              & FL   & 85.03$^*$    & 81.60       \\
       & GDL   & 81.58     & 83.80       \\
          & LL  & \textbf{85.60$^*$}     & \textbf{86.80$^*$}      \\
        & SADL & 51.10$^*$     & 56.00$^*$      \\
 \bottomrule
\end{tabular}}
    \caption{Results on Question Answering datasets. $^*$ indicates values for which $p < 0.05$.}
    \label{tab:qa_results}
\end{table*}

\section{Prompt Examples}
\label{sec:qualitative_input}

We express the prompts to fine-tune the LLMs considered as follows:\\
\textit{Question: [Question Text] (Context: [Context text]) Answer: [Answer Text]}\\
where \textit{Context} is optional as not every dataset includes it. The answer format can be either a single letter corresponding to the answer for QA or a series of passages and a final answer for mathematical reasoning problems. In the latter case, we adhere to the format of GSM8K:\\\textit{<<[Formula]>> ... \#\#\#\# [Final answer]}\\where each \textit{Formula} comprises operators and operands, which can be numbers or symbols.
This is done to better evaluate mathematical steps, which exhibit less ambiguity and adhere to stricter lexical rules than textual reasoning. In the following, we include some example prompts.

\paragraph{GSM8K}
\noindent \textit{Question: John takes care of 10 dogs.  Each dog takes .5 hours a day to walk and take care of their business.  How many hours a week does he spend taking care of dogs? \\ Answer: <<10*.5=5>> <<5*7=35>> \#\#\#\# 35}

\paragraph{MathQA}
\noindent \textit{Question: Sophia finished 2/3 of a book . she calculated that she finished 90 more pages than she has yet to read . how long is her book ? \\ Answer: <<divide(n0,n1)>> <<subtract(const\_1,\#0)>> <<divide(n2,\#1)>> \#\#\#\# 270}

\paragraph{OpenBookQA}
\textit{Question: Stars are \\ 
A. warm lights that float \\
B. made out of nitrate \\
C. great balls of gas burning billions of miles away \\
D. lights in the sky \\ 
Context: a star is made of gases \\ 
Answer: C}

\paragraph{HellaSwag}
\noindent \textit{Question: A female chef in white uniform shows a stack of baking pans in a large kitchen presenting them. the pans \\
A. contain egg yolks and baking soda. \\
B. are then sprinkled with brown sugar. \\
C. are placed in a strainer on the counter. \\
D. are filled with pastries and loaded into the oven. \\
Answer: D}

\end{document}